\documentclass[runningheads]{llncs}
\usepackage[T1]{fontenc}
\usepackage{graphicx,verbatim}
\usepackage{booktabs,multicol,multirow}
\usepackage{amsmath}
\usepackage{bbding}
\usepackage{hyperref}
\usepackage{array}
\usepackage{caption}
\usepackage{subcaption}
\usepackage{url}
\usepackage{threeparttable}
\usepackage[T1]{fontenc} 
\usepackage{wrapfig}
\begin{document}
\title{RadEyeVideo: Enhancing general-domain Large Vision Language Model for chest X-ray analysis with video representations of eye gaze}
\titlerunning{RadEyeVideo}
%

\author{Yunsoo Kim, Jinge Wu, Honghan Wu}
\authorrunning{Kim et al.}
\institute{Institute of Health Informatics\\
University College London\\
London, UK \\
\email{\{yunsoo.kim.23,jinge.wu.20,honghan.wu\}@ucl.ac.uk} \\
}

\maketitle              
\begin{abstract}
Large Vision-Language Models (LVLMs) have demonstrated promising performance in chest X-ray (CXR) analysis. To enhance human-computer interaction, several studies have incorporated radiologists' eye gaze, typically through heatmaps or textual prompts. However, these methods often overlook the sequential order of eye movements, which could provide valuable insights by highlighting both the areas of interest and the order in which they are examined. In this work, we propose a novel approach called \textbf{RadEyeVideo} that integrates radiologists’ eye-fixation data as a video sequence, capturing both the temporal and spatial dynamics of their gaze. We evaluate this method in CXR report generation and disease diagnosis using three general-domain, open-source LVLMs with video input capabilities. When prompted with eye-gaze videos, model performance improves by up to 24.6\% in the report generation task and on average 15.2\% for both tasks using scaled evaluation metrics. Notably, \textbf{RadEyeVideo} enhanced an open-domain LVLM model, LLaVA-OneVision, to surpass task-specific medical LVLMs such as MAIRA-2 and CheXagent, trained on large Chest X-ray data. This work highlights that domain expert's knowledge (eye-gaze information in this case), when effectively integrated with LVLMs, can significantly enhance general-domain models' capabilities in clinical tasks. RadEyeVideo is a step toward a scalable human-centered approach of utilizing LVLMs in medical image analytics.


\end{abstract}

\section{Introduction}
Large Vision-Language Models (LVLMs) have emerged as a potentially transformative approach for various chest X-ray (CXR) analysis, including visual question answering, automated report generation, and error detection within the reports \cite{bannur2024maira,hyland2023maira,li2024llavamed,saab2024capabilities,wu2023exploring,wu2024slava}. While these models offer the potential to streamline clinical workflows and enhance diagnostic efficiency, the reliability of LVLMs in real-world clinical environments remains a challenge. A key limitation is the variability and accuracy of the outputs generated by these models \cite{alsaad2024multimodal,chen2024detecting,wu2024hallucination,xiao2024comprehensive}.

A promising solution to enhance LVLMs in clinical contexts is human-in-the-loop integration, particularly via expert cues. Recent work demonstrates that combining human expertise with automated analysis can exceed the performance of either radiologists or AI models working alone \cite{calisto2022hitlbreast,hitlperformance}. One such expert cue is eye-tracking data, which captures both where and how long radiologists look at relevant regions. Several studies have demonstrated that integrating eye-gaze information into AI models enhances diagnostic accuracy by providing insights into the areas radiologists focus on during image interpretation \cite{ji2023hitlmammo,ma2023hitleye,wang2024gazegnn,wang2022follow,zhao2024mining}.

Recent work has expanded the use of radiologists' eye-tracking data in LVLMs for multimodal tasks like report generation and visual question answering \cite{kim2024enhancing,kim2024human}. These models typically integrate eye-gaze information through either simplified static textual prompts or heatmaps. However, these recent studies generally overlook the sequential order of eye movements, which could provide valuable additional context. The sequence in which a radiologist scans an image offers insight into how they prioritize different regions, potentially contributing to more nuanced interpretations and improving performance on downstream tasks.

\begin{figure}[h]
\begin{center}
\fbox{\includegraphics[width=.98\linewidth]{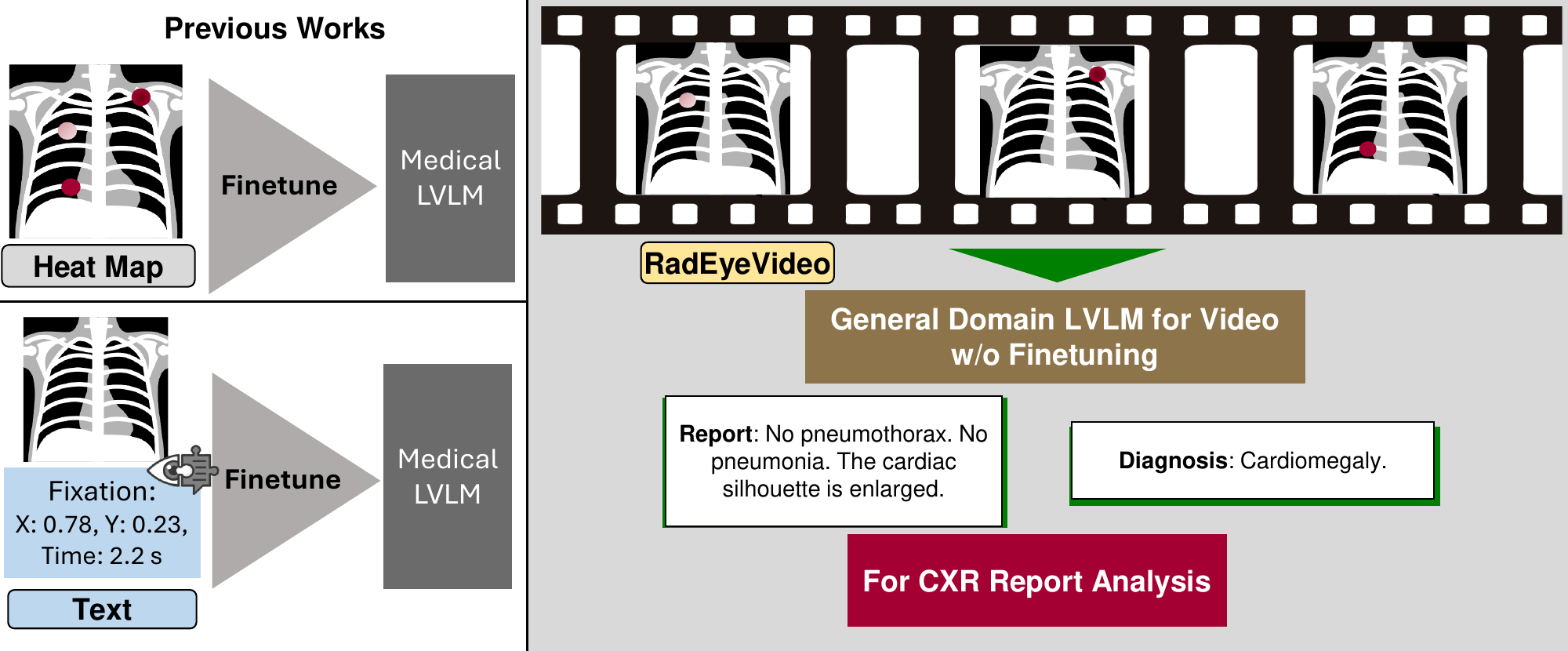}}
\end{center}
    \caption{Comparison of the eye gaze prompting methods. \textbf{Heat Map:} previous work that used the static eye gaze information laid over the chest X-ray image. \textbf{Fixation Text:} previous work that implemented eye gaze information as textual prompt in the order of duration. \textbf{RadEyeVideo:} Our prompting method implements a video to capture a dynamic representation of eye gaze.}
    \label{fig:figure1}
\end{figure}

To address this gap, we propose a novel prompting approach, \textbf{RadEyeVideo}, that integrates eye-tracking data as a video, capturing both the temporal and spatial order of eye movements (Figure \ref{fig:figure1}). This richer representation preserves the dynamic process of how radiologists navigate and prioritize different regions in an image. By incorporating this sequential flow, our approach offers deeper insights into the decision-making process, highlighting both the areas of interest and the order in which they are examined. Since radiologists often follow a structured approach during interpretation, capturing this sequence provides critical context for the model’s understanding. 

\subsubsection{Related Works}
Figure \ref{fig:figure1} shows how our approach distinguishes itself from previous prompting methods for LVLMs in CXR analysis. 

For ``Heat Map'' method, red dots are overlayed on the CXR images to visualize spatial patterns of eye gaze information \cite{kim2024enhancing}. These dots indicate the areas of focus, with darker dots representing longer gaze durations, thereby incorporating a temporal dimension to the data. However, it fails to convey any information about their sequential order, making the prompting method a simplified snapshot of gaze movements. 

Although the textual prompt can contain some sequential temporal information regarding fixations, existing methods do not fully leverage this capability; instead, gaze data is typically organized by duration, limiting its representation of sequential order \cite{kim2024human}. Although the textual format has the potential to convey a correct sequence, it ultimately represents all spatial and temporal information in plain text. The X and Y coordinates, represented in relative width and height, offer only an indirect representation. The relative position of the gaze in the text has to be used to infer the gaze point on the image, and LVLMs capability of doing this remains unexplored. 

In contrast, RadEyeVideo captures the dynamic nature of gaze patterns, providing a more comprehensive understanding of radiologist behavior during image interpretation.

\subsubsection{Contributions}
This work aims to advance human-centered AI research in AI-assisted diagnostics through innovative integration of eye-tracking data. It establishes a generalizable prompting strategy with a video that can be applied across various medical and non-medical fields.

The contributions offer significant improvements in both the accuracy of report generation and diagnosis and the potential for broader human-AI collaboration. 

This multimodal expert-in-the-loop approach aims to advance human-centered AI research in medical image computing by effectively combining artificial and human intelligence. The proposed work will evaluate the impact of radiologists' expertise or perceptual cues in the form of eye gaze on model performance, aiming to enhance the accuracy and clinical relevance of AI-driven solutions for various medical image analysis tasks. Our approach offers several key contributions to the fields of AI-driven diagnostics and human-AI collaboration, summarized as follows:


\begin{itemize}
\item \textbf{RadEyeVideo - Dynamic Eye Gaze with Video for CXR Analysis:} 
A novel prompting method that inputs radiologists' eye-gaze as short video sequences, capturing the spatial and temporal structure of scan paths. This prompting improved by 24.6\% in the report generation and 15.2\% in the overall performance, outperforming even certain task-specific medical LVLMs.

\item \textbf{Comprehensive Benchmark of Eye-Gaze Prompting:}
We systematically compare our video-based approach against heatmap-based and textual-fixation-based prompts. This is the first holistic evaluation of multiple gaze-integration methods, demonstrating RadEyeVideo’s clear advantages in diagnostic accuracy and clinical relevance.

\item \textbf{MIMIC-Eye-Video Dataset:}
We introduce a novel dataset derivation pipeline, yielding eye-gaze videos for 2,298 CXR images from MIMIC-CXR. Due to policy restrictions, only the code and protocols are shared, but this enables future research to replicate our pipeline and further advance medical AI with expert gaze data.
\end{itemize}

\section{RadEyeVideo - Dynamic Eye Gaze Video Prompting}
\label{methods}

We introduce RadEyeVideo, a dynamic eye-gaze video prompting technique, designed to incorporate radiologists' eye-gaze patterns into LVLMs for chest X-ray report generation and diagnosis. The key motivation behind this method is to enhance the interpretative capabilities of general-purpose LVLMs by leveraging human perceptual cues, such as eye-tracking data. The rationale for incorporating both spatial and temporal aspects of radiologists' gaze patterns is based on the understanding that expert radiologists do not analyze medical images in a static manner; rather, they focus on diagnostically relevant regions over time. Capturing these dynamics allows the model to gain insights from the way experts interact with medical images, thus guiding it toward better diagnostic decisions.

\subsection{Why Eye-Gaze Data?}
Radiologists demonstrate high proficiency in interpreting CXRs due to their ability to efficiently scan the image and focus on areas that are clinically significant. Eye-tracking data captures these visual search patterns and provides a rich source of information, reflecting the expert's decision-making process. RadEyeVideo translates these patterns into video-based prompts, which combine spatial attention (where radiologists look) with temporal sequencing (how long and when they look at certain regions), creating a multi-modal input that better aligns with clinical reasoning.

\subsection{Constructing the Gaze-Based Video Prompt}

Let $G = \{ g_1, g_2, \dots, g_n \}$ denote the sequence of gaze fixations for a CXR image, where each gaze fixation $ g_i = (x_i, y_i, t_i) $ consists of the spatial coordinates $ (x_i, y_i) $ and fixation duration $ t_i $. The gaze radius size was fixed to 5 pixels.



We then construct the video by representing each fixation $g_i $ over a series of frames, with the number of frames $ F_i $ proportional to the duration $ t_i $:

\begin{equation}
F_{\text{total}} = \sum_{g_i \in G} F_i = \sum_{g_i \in G} (t_i \times 10) \quad \text{where} \quad F_i = t_i \times \text{fps}
\end{equation}

where $ \text{fps} $ is the frame rate, set to 10 frames per second. 

Each frame $ v_j $ in the video sequence $ V = \{ v_1, v_2, \dots, v_{F_{\text{total}}} \} $ represents the CXR image with a red dot at coordinates $(x_i, y_i) $, indicating the radiologist's gaze position. The duration of each fixation controls the number of frames in which the gaze remains in a given position.

To make the video suitable for LVLMs, which may require a fixed number of input frames (e.g., 16), we employ a uniform sampling strategy. Instead of using all frames, we evenly sample $k$ frames from the total video sequence $F_{\text{total}}$. The index of each sampled frame is given by:

\begin{equation}
V_{\text{sampled}} = \{ v_{j} \mid j = 1, 2, \ldots, k \} \quad \text{where} \quad v_j = \left\lfloor \frac{j \cdot F_{\text{total}}}{k} \right\rfloor \quad \text{for} \quad j = 1, 2, \ldots, k
\end{equation}

where $k$ is the number of frames to sample, ensuring a balanced representation of the gaze data.

This sampling process, combined with the weighted number of frames based on duration, effectively captures important gaze patterns while reducing computational overhead. This approach maintains the temporal distribution and sequential order of gaze fixations, ensuring that key insights are preserved.

\subsection{Input Representation}

The input to the LVLM consists of two components: a textual prompt $ T $ and the generated video sequence $ V $. The textual prompt specifies the task (e.g., "What are the possible differential diagnoses for this patient?"), while the video provides spatio-temporal information about the radiologist's eye movements. 

\subsection{Computational Efficiency of RadEyeVideo}
The incorporation of eye gaze data as 
a video, rather than as a heatmap, results in only a marginal increase in GPU memory usage—approximately 2GB, rising from 20,076MB (eye gaze heatmap) to 21,954MB (eye gaze video) when using the LLaVA-OneVision model. This corresponds to a modest 10\% increase in GPU memory consumption.

Notably, this increase is minimal when compared to the significantly higher computational demands of training specialized medical LVLMs, which often require at least double the computational resources. The efficiency of our approach underscores its practicality for deployment, even in resource-constrained environments, while achieving enhanced diagnostic performance.

\subsection{Video Prompting for Report Generation}

In clinical practice, radiologists' reports serve as crucial foundations for diagnostic and treatment decisions. CXR reports typically comprise two main sections: ``Findings" and ``Impressions." The ``Findings" section meticulously details the radiologist's observations from the images, requiring keen observation skills and specialized knowledge. The ``Impressions" section, on the other hand, provides a concise summary of the ``Findings", offering clinicians a quick yet accurate diagnostic reference. In this case the report generation task can be split into two subtasks: \textbf{Findings generation} and \textbf{Impression generation}. Mathematically speaking, 
\begin{equation}
Y_F = f_F(T, V, E)
\end{equation}
\begin{equation}
Y_I = f_I(T, F, V, E)
\end{equation}
where: $Y_F$ denotes the output of the Findings generation task, $Y_I$ denotes the output of the Impression summarization task, $T$ denotes the task instructions or prompts for each task, $V$ represents the sampled video prompts. To align AI-generated reports more closely with authentic clinical reporting styles, we provide three exemplar reports, $E$, as in-context learning. In the Impression geneartion task, $F$ denotes the Findings from original reports, which serves as an additional input for the Impression task. 

\subsection{Video Prompting for Diagnosis}

Diagnosis is a critical component of chest radiograph (CXR) interpretation, where radiologists synthesize observed abnormalities to form a conclusive diagnostic assessment. This process demands not only acute visual perception but also extensive clinical knowledge and reasoning skills. In our study, we adapt our video-based eye-gaze prompting methodology to assist Large Vision-Language Models (LVLMs) in generating more accurate and clinically relevant diagnoses. The diagnostic task can be mathematically represented as:
\begin{equation}
Y_D = f_D(T, V, E)
\end{equation}
where $Y_D$ denotes the output of the diagnostic task.

\section{Experiment}

\subsection{Dataset}

\begin{table}[t]
    \centering
    \caption{Summary of images and dictated reports}
    \begin{tabular}{lcc}
        \toprule
        \textbf{Category} & \textbf{Alpha Set} & \textbf{Beta Set} \\
         \midrule 
         Images  & 3,134 & 549 \\
         Number of Radiologists per image & 1.18 & 5.82 \\
         Dictated Reports & 3,699 & 3,197 \\
    \bottomrule
    \end{tabular}
    \label{tab:dataset_stat}
\end{table}

We conduct our experiments on a combined dataset derived from \textbf{Eye Gaze} and \textbf{REFLACX} \cite{karargyris2020eye,lanfredi2021reflacx}, which together provide 3,693 unique CXR images with radiologist eye-tracking data. Since multiple radiologists can annotate a single image, the total number of \textbf{dictated reports} across images is 6,896. 

These images come from MIMIC-CXR \cite{johnson2019mimic}. However, because the dictated reports here differ from the original MIMIC-CXR textual reports, we leverage both training and test images into two evaluation subsets, labeled \textbf{alpha} and \textbf{beta}, to help monitor any potential contamination issues when evaluating performance. Table~\ref{tab:dataset_stat} summarizes the dataset statistics, including image counts and average radiologists per image.

\subsection{Models}

\begin{table}[htbp]
\footnotesize
\centering
    \caption{Model descriptions. Training features are abbreviated as follows: Report Generation (R) and Diagnosis (D). Models in bold are trained with the MIMIC-CXR dataset.}
    \label{tab:overview_models}
\begin{tabular}{@{}lccccccc@{}}
\toprule
  \multirow{2}{*}[-3pt]{\textbf{Model Name}}  & \multirow{2}{*}[-3pt]{\textbf{Size}} & \multirow{2}{*}[-3pt]{\textbf{Backbone LLM}} & \multicolumn{2}{c}{\bf Trained Task} & \multicolumn{3}{c}{\bf Supported Modalities}  \\ \cmidrule(lr){4-5}   \cmidrule(lr){6-8}     
  & & &    \textbf{R} &  \textbf{D} & \textbf{Image} &  \textbf{Text} &  \textbf{Video} \\
\midrule
\textbf{CheXagent}\cite{chen2024chexagent} & 8B & Mistral 7B & \Checkmark & \Checkmark  & \Checkmark  & \Checkmark  & \XSolidBrush \\
\textbf{CXR-LLaVA}\cite{lee2023cxr} & 8B & LLaMA2 7B & \Checkmark  & \Checkmark & \Checkmark  & \Checkmark  & \XSolidBrush \\
\textbf{MAIRA-2}\cite{bannur2024maira} & 8B & Vicuna v1.5 7B & \Checkmark  & \Checkmark & \Checkmark  & \Checkmark  & \XSolidBrush \\
 \midrule 
LongVA\cite{zhang2024long} & 8B & Qwen2 7B & - & - & \Checkmark  & \Checkmark  & \Checkmark  \\
VideoLLaMA3\cite{zhang2025videollama} & 8B & Qwen2.5 7B & - & - & \Checkmark  & \Checkmark  & \Checkmark  \\
LLaVA-OneVision\cite{li2024llavaov} & 8B & Qwen2 7B & - & - & \Checkmark  & \Checkmark  & \Checkmark  \\
\bottomrule
\end{tabular}

\end{table}

In this study, we selected a range of models to investigate various prompting methods for integrating eye gaze information into the CXR report generation process. Specifically, we focused on LongVA-7B, VideoLLaMA3-8B, and LLaVA-OneVision-7B, which are the latest LVLMs that can handle videos  \cite{li2024llavaov,zhang2025videollama,zhang2024long}. 

For baseline comparisons, we included CXR LVLMs such as CXR-LLaVA, CheXagent, MAIRA-2 \cite{bannur2024maira,chen2024chexagent,lee2023cxr}. These models have been trained with the entire MIMIC-CXR training split which is about 200,000 CXR images and reports for diverse tasks, including report generation. 



\subsection{Evaluation}

\begin{figure}[t]
    \centering
    \fbox{\includegraphics[width=0.98\linewidth]{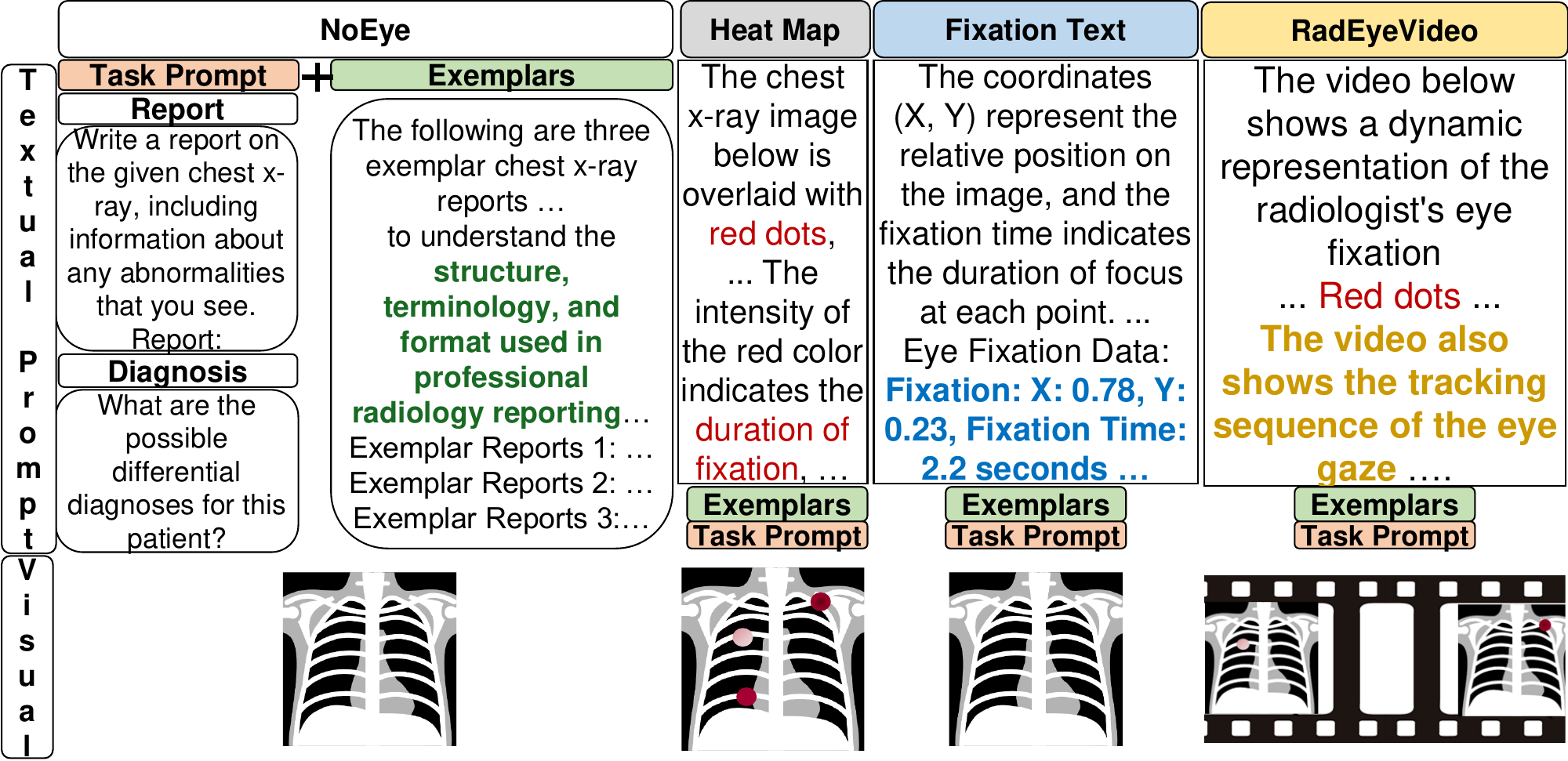}}
    \caption{Eye gaze textual and visual prompts. Texts are highlighted in different colors to emphasize the important aspect of each prompting method.}
    \label{fig:prompt}
\end{figure}

In addition to using eye-tracking data as video sequences, we explore other methods of integrating eye-gaze information into LVLMs for CXR report analysis, including heatmaps and textual prompts. We systematically evaluate these methods to determine which most effectively enhances report generation and diagnosis, focusing on accuracy, completeness, and clinical relevance. Figure \ref{fig:prompt} provides a comprehensive overview of the various textual and visual prompts used to enhance report generation and diagnosis for chest X-rays.


\subsubsection{Evaluation Hyperparameters}
\label{eval_setting}
For evaluation, we used a batch size of 1 and a temperature of 0. A temperature of 0 was chosen to minimize the randomness in the generated text produced by the model. When the temperature of 0 is not accepted by the model, such as the VideoLLaMA3 model, then we used 0.1 instead. The maximum length of the model's responses for each task was set as 256 for report generation and 64 for diagnosis. This setup ensured consistent and efficient experimentation across both tasks.
 
\subsubsection{Evaluation Metric}
We evaluated the generated reports using a combination of radiology-specific metrics. We used CheXbert (micro F1 for top 5 abnormalities), RadGraph-XL, and RaTEScore \cite{delbrouck2024radgraphxl,smit2020chexbert,zhao2024ratescore}. 

To compare the performance of our models with respect to both general lexical and radiology-specific metrics, we introduce a scaling based on CheXagent, a medical LVLM that is trained for both report generation and diagnosis and known to perform the best among the LVLMs for these two tasks \cite{chen2024chexagent}. 

Let \( S_{m,i} \) represent the raw score for metric \( m \) on model \( i \), and let \( S_{m,\textit{CheXagent}} \) represent the score of another model for the same metric \( m \). The scaled score \( \hat{S}_{m,i} \) for each model \( i \) on metric \( m \) and the average of the scaled scores across all metrics \( M\) for the the overall performance score, \( \hat{S}_i \), are then calculated as follows:

\begin{equation}
\hat{S}_{m,i} = \frac{S_{m,i}}{S_{m,\textit{CheXagent}}} \times 100, \qquad \hat{S}_i = \frac{1}{|M|} \sum_{m \in M} \hat{S}_{m,i}
\end{equation}

This scaling ensures that all metrics are directly comparable and highlights relative improvements (or declines) across models, regardless of the absolute values.

\section{Results and Discussion}
\label{sec:results}

\begin{table}[h]
\footnotesize
\centering
\caption{Eye gaze evaluation results. 
NoEye prompting scores are reported with model names. Bold - best scores (excluding scores from MIMIC-CXR trained models); parentheses -  performance improvements.} 
\begin{tabular}{@{}lcccccc@{}}

\toprule
  \multirow{2}{*}[-3pt]{\textbf{Methods}}  & \multicolumn{2}{c}{\bf Report} & \multicolumn{2}{c}{\bf Diagnosis} & \multirow{2}{*}[-3pt]{\textbf{Overall}} \\ \cmidrule(lr){2-3}   \cmidrule(lr){4-5}  
  & \textbf{Alpha} & \textbf{Beta}    & \textbf{Alpha} & \textbf{Beta} & & \\
\midrule
CheXagent
 & 100.0 &	100.0 &	100.0 &	100.0 &	100.0 \\
CXR-LLaVA  & 374.7 &	311.3 &		62.2 &	62.7 &	202.7 \\
MAIRA-2  & 234.4 &	251.1 &		68.2 &	60.9 &	152.2 \\
\midrule
VideoLLaMA3 & 267.3 &	\textbf{245.2} &	48.7 &	44.0 &	151.3 \\
\hspace{1em} w/ Heat Map & 206.7 &	180.3 &	53.9 &	50.1 &	122.7 \\
\hspace{1em} w/ Fixation Text & 245.8 &	228.9 &	50.2 &	45.9 &	142.7 \\
\hspace{1em} w/ RadEyeVideo (Ours) & 
197.2 &	211.9 &	56.9 &	58.4 &	131.1 \\
& (-70.1) & (-33.3) & \textbf{(+8.2)} & \textbf{(+14.4)} & (-20.2) \\
\midrule
LongVA & 222.0 &	203.3 &	53.9 &	54.6 &	133.5 \\
\hspace{1em} w/ Heat Map & 165.6 &	159.7 &	\textbf{66.9} &	\textbf{61.7} &	113.5	 \\
\hspace{1em} w/ Fixation Text & 175.1 &	165.1 &	56.9 &	54.1 &	112.8	 \\
\hspace{1em} w/ RadEyeVideo (Ours) & 233.2 &	199.5 &	54.3 &	55.3 &	135.6 \\
 & (+10.2) & (-3.8) & (+0.4) & (+0.7) & (+2.1) \\
\midrule
LLaVA-OneVision & 237.1 &	216.1 &	52.2 &	52.4 &	139.4 \\
\hspace{1em} w/ Heat Map & 214.4 &	210.5 &	51.1 &	51.0 &	131.8 \\
\hspace{1em} w/ Fixation Text & 214.7 &	215.0 &	52.2 &	53.6 &	133.9 \\
\hspace{1em} w/ RadEyeVideo (Ours) & \textbf{269.5} &	232.9 &	57.3 &	58.5 &	\textbf{154.6} \\
& \textbf{(+32.4)} & \textbf{(+16.8)} & (+5.1) & (+6.1) & \textbf{(+15.2)}  \\
\bottomrule
\end{tabular}
\label{tab:eye_gaze}
\end{table}

The evaluation results in Table \ref{tab:eye_gaze} offer a comprehensive overview of performance across two tasks, report generation and diagnosis, for both the alpha and beta splits.

\subsubsection{RadEyeVideo Empowers LLaVA-OneVision to Surpass Specialized CXR Models}
Among the general-domain LVLMs we evaluated, \textbf{LLaVA-OneVision with RadEyeVideo} emerges as the top performer overall, reaching an average scaled score of \textbf{154.6\%}, which exceeds both \textbf{CheXagent} and \textbf{MAIRA-2}. Notably, on at least one of the evaluated splits, LLaVA-OneVision with RadEyeVideo outperforms these specialized models in report generation, illustrating how spatiotemporal gaze prompts can bridge domain gaps without exhaustive medical pretraining. This result highlights that a sufficiently flexible LVLM can leverage expert eye fixation sequences in a video format to match or surpass specialized CXR-trained counterparts on critical clinical tasks.

\subsubsection{RadEyeVideo Effective in Diagnosis but Mixed for Report Generation}
All three general-domain LVLMs see a boost in diagnosis after integrating RadEyeVideo, yet exhibit mixed results in report generation. \textbf{LLaVA-OneVision} is the only model that increased performance for both tasks. VideoLLaMA3 gains +8.2\% (alpha) and +14.4\% (beta) for diagnosis—the largest improvement among general models—but suffers the largest decrease in report scores, -70.1\% (alpha) and -33.3\% (beta). LongVA achieves slight diagnostic gains (+0.4 alpha, +0.7 beta) and an increase only in the alpha split for the report generation by +10.2\%. This discrepancy underscores how architecture-specific factors such as the backbone large language model may determine whether a model can effectively synthesize gaze-driven video prompts without compromising textual coherence.

\subsubsection{Limitations of Heatmap and Fixation Text Prompts}

\begin{figure}[htbp!]
    \centering
    \fbox{\includegraphics[trim={0cm 0cm 0cm 0cm},clip,width=0.98\linewidth]{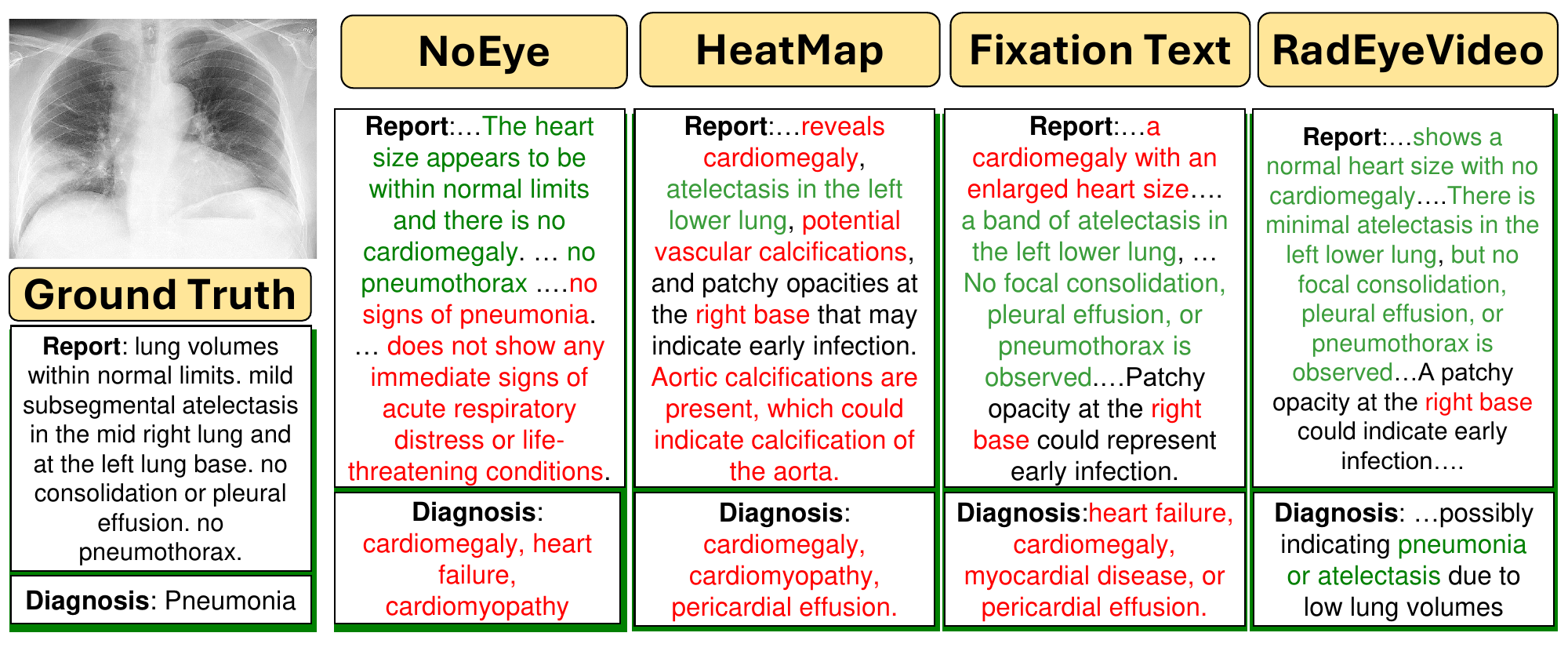}}
    \caption{Sample response from LLaVA-OneVision. \textbf{Green} - correct statements about the CXR; \textbf{Red} - incorrect statements. To comply with the MIMIC-CXR data usage license, the CXR image is substituted with a Wikimedia image depicting the same disease, and the text report is paraphrased.}
    \label{fig:figure3}
\end{figure}

While RadEyeVideo shows strong results, both Heat Map prompting and Fixation Text prompting methods do not always improve performance. While all eye gaze methods improve performance in diagnosis, only \textbf{RadEyeVideo} showed improvement in report generation of LongVA for the alpha split and LLaVA-OneVision for the beta split. This result suggests that simply highlighting fixation intensities (Heat Map) or listing them in textual form (Fixation Text) neither captures the full spatio-temporal flow of expert scanning nor integrates well with LVLMs for longer output.

Figure \ref{fig:figure3} exemplifies this improvement in both the report generation and diagnosis. While the generated reports showed slight improvements with more medical concepts correctly stated, errors like the incorrect location of an abnormality remain, indicating room for further improvement. Figure \ref{fig:figure3} also shows that only RadEyeVideo empowered the model to correctly predict pneumonia.

\section{Conclusion}

In this work, we introduced RadEyeVideo, a novel video prompting approach, into general domain LVLMs, enhancing their performance in chest X-ray report generation and diagnosis. Our approach effectively captures the spatial and temporal dynamics of eye gaze, outperforming existing eye-gaze prompting techniques and even the task-specific medical LVLMs. This method demonstrates significant potential for bridging the gap between general-purpose and task-specific models in medical imaging. Future work will explore extending video-based prompting to tasks like visual question answering and anatomical structure detection. Additionally, applying this method to other medical imaging areas, such as CT and MRI scans, holds promise for improving accuracy and clinical relevance across various domains in healthcare.

\subsubsection*{Limitation}

This study's evaluation was limited by the small dataset size, due to the difficulty of obtaining radiologists' eye-tracking data. Furthermore, the MIMIC-CXR dataset may not fully capture the diversity of real-world medical imaging. Future work should focus on larger and more diverse datasets to better assess the method's generalizability across different imaging modalities.

\subsubsection*{Ethical Statement}
This research adhered to the data usage agreements of the MIMIC-EYE dataset and maintained strict compliance with privacy regulations.
All data used were de-identified, ensuring ethical standards were met throughout the study.

\bibliographystyle{splncs04}
\bibliography{MICCAI2025}

\begin{thebibliography}{10}
\providecommand{\url}[1]{\texttt{#1}}
\providecommand{\urlprefix}{URL }
\providecommand{\doi}[1]{https://doi.org/#1}

\bibitem{alsaad2024multimodal}
AlSaad, R., Abd-Alrazaq, A., Boughorbel, S., Ahmed, A., Renault, M.A., Damseh, R., Sheikh, J.: Multimodal large language models in health care: Applications, challenges, and future outlook. Journal of Medical Internet Research  \textbf{26},  e59505 (2024)

\bibitem{bannur2024maira}
Bannur, S., Bouzid, K., Castro, D.C., Schwaighofer, A., Bond-Taylor, S., Ilse, M., P{\'e}rez-Garc{\'\i}a, F., Salvatelli, V., Sharma, H., Meissen, F., et~al.: Maira-2: Grounded radiology report generation. arXiv preprint arXiv:2406.04449  (2024)

\bibitem{calisto2022hitlbreast}
Calisto, F.M., Santiago, C., Nunes, N., Nascimento, J.C.: Breastscreening-ai: Evaluating medical intelligent agents for human-ai interactions. Artificial Intelligence in Medicine  \textbf{127},  102285 (2022)

\bibitem{chen2024detecting}
Chen, J., Yang, D., Wu, T., Jiang, Y., Hou, X., Li, M., Wang, S., Xiao, D., Li, K., Zhang, L.: Detecting and evaluating medical hallucinations in large vision language models. arXiv preprint arXiv:2406.10185  (2024)

\bibitem{chen2024chexagent}
Chen, Z., Varma, M., Delbrouck, J.B., Paschali, M., Blankemeier, L., Van~Veen, D., Valanarasu, J.M.J., Youssef, A., Cohen, J.P., Reis, E.P., et~al.: Chexagent: Towards a foundation model for chest x-ray interpretation. arXiv preprint arXiv:2401.12208  (2024)

\bibitem{delbrouck2024radgraphxl}
Delbrouck, J.B., Chambon, P., Chen, Z., Varma, M., Johnston, A., Blankemeier, L., Van~Veen, D., Bui, T., Truong, S., Langlotz, C.: Radgraph-xl: A large-scale expert-annotated dataset for entity and relation extraction from radiology reports. In: Findings of the Association for Computational Linguistics ACL 2024. pp. 12902--12915 (2024)

\bibitem{hyland2023maira}
Hyland, S.L., Bannur, S., Bouzid, K., Castro, D.C., Ranjit, M., Schwaighofer, A., P{\'e}rez-Garc{\'\i}a, F., Salvatelli, V., Srivastav, S., Thieme, A., et~al.: Maira-1: A specialised large multimodal model for radiology report generation. arXiv preprint arXiv:2311.13668  (2023)

\bibitem{ji2023hitlmammo}
Ji, C., Du, C., Zhang, Q., Wang, S., Ma, C., Xie, J., Zhou, Y., He, H., Shen, D.: Mammo-net: Integrating gaze supervision and interactive information in multi-view mammogram classification. In: International Conference on Medical Image Computing and Computer-Assisted Intervention. pp. 68--78. Springer (2023)

\bibitem{johnson2019mimic}
Johnson, A.E., Pollard, T.J., Berkowitz, S.J., Greenbaum, N.R., Lungren, M.P., Deng, C.y., Mark, R.G., Horng, S.: Mimic-cxr, a de-identified publicly available database of chest radiographs with free-text reports. Scientific data  \textbf{6}(1), ~317 (2019)

\bibitem{karargyris2020eye}
Karargyris, A., Kashyap, S., Lourentzou, I., Wu, J., Tong, M., Sharma, A., Abedin, S., Beymer, D., Mukherjee, V., Krupinski, E., et~al.: Eye gaze data for chest x-rays. PhysioNet https://doi. org/10.13026/QFDZ-ZR67  (2020)

\bibitem{lanfredi2021reflacx}
Lanfredi, R.B., Zhang, M., Auffermann, W., Chan, J., Duong, P.A., Srikumar, V., Drew, T., Schroeder, J., Tasdizen, T.: Reflacx: Reports and eye-tracking data for localization of abnormalities in chest x-rays (2021)

\bibitem{lee2023cxr}
Lee, S., Youn, J., Kim, M., Yoon, S.H.: Cxr-llava: Multimodal large language model for interpreting chest x-ray images. arXiv preprint arXiv:2310.18341  (2023)

\bibitem{li2024llavaov}
Li, B., Zhang, Y., Guo, D., Zhang, R., Li, F., Zhang, H., Zhang, K., Zhang, P., Li, Y., Liu, Z., et~al.: Llava-onevision: Easy visual task transfer. arXiv preprint arXiv:2408.03326  (2024)

\bibitem{li2024llavamed}
Li, C., Wong, C., Zhang, S., Usuyama, N., Liu, H., Yang, J., Naumann, T., Poon, H., Gao, J.: Llava-med: Training a large language-and-vision assistant for biomedicine in one day. Advances in Neural Information Processing Systems  \textbf{36} (2024)

\bibitem{ma2023hitleye}
Ma, C., Zhao, L., Chen, Y., Wang, S., Guo, L., Zhang, T., Shen, D., Jiang, X., Liu, T.: Eye-gaze-guided vision transformer for rectifying shortcut learning. IEEE Transactions on Medical Imaging  (2023)

\bibitem{hitlperformance}
Patel, B.N., Rosenberg, L., Willcox, G., Baltaxe, D., Lyons, M., Irvin, J., Rajpurkar, P., Amrhein, T., Gupta, R., Halabi, S., Langlotz, C., Lo, E., Mammarappallil, J., Mariano, A.J., Riley, G., Seekins, J., Shen, L., Zucker, E., Lungren, M.P.: Human--machine partnership with artificial intelligence for chest radiograph diagnosis. npj Digital Medicine  \textbf{2}(1), ~111 (2019). \doi{10.1038/s41746-019-0189-7}, \url{https://doi.org/10.1038/s41746-019-0189-7}

\bibitem{saab2024capabilities}
Saab, K., Tu, T., Weng, W.H., Tanno, R., Stutz, D., Wulczyn, E., Zhang, F., Strother, T., Park, C., Vedadi, E., et~al.: Capabilities of gemini models in medicine. arXiv preprint arXiv:2404.18416  (2024)

\bibitem{smit2020chexbert}
Smit, A., Jain, S., Rajpurkar, P., Pareek, A., Ng, A.Y., Lungren, M.P.: Chexbert: combining automatic labelers and expert annotations for accurate radiology report labeling using bert. arXiv preprint arXiv:2004.09167  (2020)

\bibitem{kim2024enhancing}
Springer: Enhancing human-computer interaction in chest x-ray analysis using vision and language model with eye gaze patterns (2024)

\bibitem{kim2024human}
Springer: Human-in-the-Loop Chest X-Ray Diagnosis: Enhancing Large Multimodal Models with Eye Fixation Inputs (2024)

\bibitem{wang2024gazegnn}
Wang, B., Pan, H., Aboah, A., Zhang, Z., Keles, E., Torigian, D., Turkbey, B., Krupinski, E., Udupa, J., Bagci, U.: Gazegnn: A gaze-guided graph neural network for chest x-ray classification. In: Proceedings of the IEEE/CVF Winter Conference on Applications of Computer Vision. pp. 2194--2203 (2024)

\bibitem{wang2022follow}
Wang, S., Ouyang, X., Liu, T., Wang, Q., Shen, D.: Follow my eye: Using gaze to supervise computer-aided diagnosis. IEEE Transactions on Medical Imaging  \textbf{41}(7),  1688--1698 (2022)

\bibitem{wu2023exploring}
Wu, J., Kim, Y., Keller, E.C., Chow, J., Levine, A.P., Pontikos, N., Ibrahim, Z., Taylor, P., Williams, M.C., Wu, H.: Exploring multimodal large language models for radiology report error-checking. arXiv preprint arXiv:2312.13103  (2023)

\bibitem{wu2024slava}
Wu, J., Kim, Y., Shi, D., Cliffton, D., Liu, F., Wu, H.: Slava-cxr: Small language and vision assistant for chest x-ray report automation. arXiv preprint arXiv:2409.13321  (2024)

\bibitem{wu2024hallucination}
Wu, J., Kim, Y., Wu, H.: Hallucination benchmark in medical visual question answering. arXiv preprint arXiv:2401.05827  (2024)

\bibitem{xiao2024comprehensive}
Xiao, H., Zhou, F., Liu, X., Liu, T., Li, Z., Liu, X., Huang, X.: A comprehensive survey of large language models and multimodal large language models in medicine. arXiv preprint arXiv:2405.08603  (2024)

\bibitem{zhang2025videollama}
Zhang, B., Li, K., Cheng, Z., Hu, Z., Yuan, Y., Chen, G., Leng, S., Jiang, Y., Zhang, H., Li, X., et~al.: Videollama 3: Frontier multimodal foundation models for image and video understanding. arXiv preprint arXiv:2501.13106  (2025)

\bibitem{zhang2024long}
Zhang, P., Zhang, K., Li, B., Zeng, G., Yang, J., Zhang, Y., Wang, Z., Tan, H., Li, C., Liu, Z.: Long context transfer from language to vision. arXiv preprint arXiv:2406.16852  (2024)

\bibitem{zhao2024ratescore}
Zhao, W., Wu, C., Zhang, X., Zhang, Y., Wang, Y., Xie, W.: Ratescore: A metric for radiology report generation. medRxiv pp. 2024--06 (2024)

\bibitem{zhao2024mining}
Zhao, Z., Wang, S., Wang, Q., Shen, D.: Mining gaze for contrastive learning toward computer-assisted diagnosis. In: Proceedings of the AAAI Conference on Artificial Intelligence. vol.~38, pp. 7543--7551 (2024)

\end{thebibliography}
%








\appendix
\newpage

\end{document}